\documentclass[]{interact}

\usepackage{epstopdf}
\usepackage{subfigure}
\usepackage{tabularx}
\usepackage{graphicx}
\usepackage{amsmath}
\usepackage{amssymb}
\usepackage{booktabs}
\usepackage{placeins} 
\usepackage{float} 
\usepackage{enumitem}
\usepackage{tabularx} 
\usepackage{longtable} 
\usepackage{array} 
\usepackage{ragged2e} 
\usepackage{booktabs} 
\usepackage{placeins}
\usepackage{afterpage} 
\usepackage{caption} 
\usepackage{booktabs} 
\usepackage{adjustbox} 
\usepackage{graphicx} 
\usepackage{tabularx}

\theoremstyle{plain}

\theoremstyle{definition}

\theoremstyle{remark}

\begin{document}


\title{Dynamic 3D KAN Convolution with Adaptive Grid Optimization for Hyperspectral Image Classification}

\author{
\name{Guandong Li\textsuperscript{a}\thanks{CONTACT Guandong Li. Email: leeguandon@gmail.com} and Mengxia Ye\textsuperscript{b}}
\affil{\textsuperscript{a}iFLYTEK, Shushan, Hefei, Anhui, China; \textsuperscript{b}Aegon THTF,Qinghuai,Nanjing,Jiangsu,China}
}

\maketitle

\begin{abstract}
Deep neural networks face several challenges in hyperspectral image classification, including high-dimensional data, sparse distribution of ground objects, and spectral redundancy, which often lead to classification overfitting and limited generalization capability. To more efficiently adapt to ground object distributions while extracting image features without introducing excessive parameters and skipping redundant information, this paper proposes KANet based on an improved 3D-DenseNet model, consisting of 3D KAN Conv and an adaptive grid update mechanism. By introducing learnable univariate B-spline functions on network edges, specifically by flattening three-dimensional neighborhoods into vectors and applying B-spline-parameterized nonlinear activation functions to replace the fixed linear weights of traditional 3D convolutional kernels, we precisely capture complex spectral-spatial nonlinear relationships in hyperspectral data. Simultaneously, through a dynamic grid adjustment mechanism, we adaptively update the grid point positions of B-splines based on the statistical characteristics of input data, optimizing the resolution of spline functions to match the non-uniform distribution of spectral features, significantly improving the model's accuracy in high-dimensional data modeling and parameter efficiency, effectively alleviating the curse of dimensionality. This characteristic demonstrates superior neural scaling laws compared to traditional convolutional neural networks and reduces overfitting risks in small-sample and high-noise scenarios. KANet enhances model representation capability through a 3D dynamic expert convolution system without increasing network depth or width. The proposed method demonstrates superior performance on IN, UP, and KSC datasets, outperforming mainstream hyperspectral image classification approaches.
\end{abstract}


\section{Introduction}

Hyperspectral remote sensing images (HSI) play a crucial role in spatial information applications due to their unique narrow-band imaging characteristics. The imaging equipment synchronously records both spectral and spatial position information of sampling points, integrating them into a three-dimensional data structure containing two-dimensional space and one-dimensional spectrum. As an important application of remote sensing technology, ground object classification demonstrates broad value in fields including ecological assessment, transportation planning, agricultural monitoring, land management, and geological surveys \cite{chang2003hyperspectral,bing2011intelligent}. However, hyperspectral remote sensing images face several challenges in ground object classification. First, hyperspectral data typically exhibits high-dimensional characteristics, with each pixel containing hundreds or even thousands of bands, leading to data redundancy and computational complexity while potentially causing the "curse of dimensionality," making classification models prone to overfitting under sparse sample conditions. Second, sparse ground object distribution means training samples are often limited, particularly for certain rare categories where annotation costs are high and distributions are imbalanced, further constraining model generalization capability. Additionally, hyperspectral images are frequently affected by noise, atmospheric interference, and mixed pixels, reducing signal-to-noise ratio and increasing difficulty in extracting features from sparse ground objects. Finally, ground object sparsity may also lead to insufficient spatial context information, causing loss in classification and recognition.

Deep learning methods for HSI classification \cite{li2019doubleconvpool,li2020hyperspectral,li2022faster,li2023dgcnet,li2025spatial_geometry,li20253d,li2018scene} have achieved significant progress. In \cite{lee2017going} and \cite{zhao2016spectral}, Principal Component Analysis (PCA) was first applied to reduce the dimensionality of the entire hyperspectral data, followed by extracting spatial information from neighboring regions using 2D CNN. Methods like 2D-CNN \cite{makantasis2015deep,chen2016deep} require separate extraction of spatial and spectral features, failing to fully utilize joint spatial-spectral information and necessitating complex preprocessing. \cite{wang2018fast} proposed a Fast Dense Spectral-Spatial Convolutional Network (FDSSC) based on dense networks, constructing 1D-CNN and 3D-CNN dense blocks connected in series. FSKNet \cite{li2022faster} introduced a 3D-to-2D module and selective kernel mechanism, while 3D-SE-DenseNet \cite{li2020hyperspectral} incorporated the SE mechanism into 3D-CNN to correlate feature maps between different channels, activating effective information while suppressing ineffective information in feature maps. DGCNet \cite{li2023dgcnet} designed dynamic grouped convolution (DGC) on 3D convolution kernels, where DGC introduces small feature selectors for each group to dynamically determine which part of input channels to connect based on activations of all input channels. Multiple groups can capture different complementary visual/semantic features of input images, enabling CNNs to learn rich feature representations. DHSNet \cite{liu2025dual} proposed a novel Central Feature Attention-Aware Convolution (CFAAC) module that guides attention to focus on central features crucial for capturing cross-scene invariant information. To leverage the advantages of both CNN and Transformer, many studies have attempted to combine CNN and Transformer to utilize local and global feature information of HSI. \cite{sun2022spectral} proposed a Spectral-Spatial Feature Tokenization Transformer (SSFTT) network that extracts shallow features through 3D and 2D convolutional layers and uses Gaussian-weighted feature tokens to extract high-level semantic features in the transformer encoder. In \cite{liu2024kan}, the authors demonstrated that KANs significantly outperform traditional multilayer perceptrons (MLPs) in satellite traffic prediction tasks, achieving more accurate predictions with fewer learnable parameters. In \cite{vaca2024kolmogorov}, KANs were recognized as a promising alternative for efficient image analysis in remote sensing, highlighting their effectiveness in this field.Some Transformer-based methods \cite{hong2021spectralformer,zhao2016spectral,liu2021swin} employ grouped spectral embedding and transformer encoder modules to model spectral representations, but these methods have obvious shortcomings - they treat spectral bands or spatial patches as tokens and encode all tokens, resulting in significant redundant computations. However, HSI data already contains substantial redundant information, and their accuracy often falls short compared to 3D-CNN-based methods while requiring greater computational complexity.

3D-CNN possesses the capability to sample simultaneously in both spatial and spectral dimensions, maintaining the spatial feature extraction ability of 2D convolution while ensuring effective spectral feature extraction. 3D-CNN can directly process high-dimensional data, eliminating the need for preliminary dimensionality reduction of hyperspectral images. However, the 3D-CNN paradigm has significant limitations - when simultaneously extracting spatial and spectral features, it may incorporate irrelevant or inefficient spatial-spectral combinations into computations. For instance, certain spatial features may be prominent in specific bands while appearing as noise or irrelevant information in other bands, yet 3D convolution still forcibly combines these low-value features. Computational and dimensional redundancy can easily trigger overfitting risks, further limiting model generalization capability. Currently widely used methods such as DFAN \cite{zhang2020deep}, MSDN \cite{zhang2019multi}, 3D-DenseNet \cite{zhang2019three}, and 3D-SE-DenseNet\cite{li2020hyperspectral} employ operations like dense connections. While dense connections directly link each layer to all its preceding layers, enabling feature reuse, they introduce redundancy when subsequent layers do not require early features. Therefore, how to more efficiently enhance the representational capability of 3D convolution kernels in 3D convolution, achieving more effective feature extraction with fewer cascaded 3D convolution kernels and dense connections while optimizing the screening and skipping of redundant information has become a direction in hyperspectral classification.

This work is based on \cite{liu2024kan}. KANs avoid the curse of dimensionality by decomposing high-dimensional functions into combinations of multiple one-dimensional functions. Specifically, the Kolmogorov-Arnold representation theorem shows that a multivariate continuous function can be represented as a finite composition of univariate continuous functions and binary addition operations. KANs utilize this property to decompose high-dimensional functions into sets of one-dimensional functions and approximate high-dimensional functions by learning combinations of these one-dimensional functions. This method not only reduces model complexity but also improves approximation capability and computational efficiency. KANet proposes a groundbreaking 3D Kolmogorov-Arnold convolution module that replaces traditional 3D convolution kernels' fixed linear weights with learnable univariate B-spline functions on network edges, significantly enhancing the modeling capability for complex spectral-spatial nonlinear relationships in hyperspectral data. By dynamically adjusting spline grid points to adapt to data distribution, it further improves model accuracy and parameter efficiency. The core transformation can be expressed as $\mathbf{y} = \sum_{i} \phi_{i}(\mathbf{x}_i; \mathbf{c}_i, \mathbf{t}_i)$, where $\phi_{i}$ represents the nonlinear activation function parameterized by B-spline basis functions, $\mathbf{c}_i$ are trainable spline coefficients, and $\mathbf{t}_i$ are adaptive grid points. The grid is dynamically adjusted to fit the non-uniform spectral distribution of hyperspectral data through the optimization objective $\min_{\mathbf{c}_i, \mathbf{t}_i} \|\mathbf{y} - \hat{\mathbf{y}}\|_2^2$.Compared with traditional 3D convolution (e.g., the fixed-weight linear transformation $\mathbf{y} = \mathbf{W} * \mathbf{x} + \mathbf{b}$), this module not only preserves the spatial locality of convolution operations but also achieves fine-grained modeling of high-dimensional spectral features through learnable nonlinear functions $\phi_{i}$, overcoming the limitations of traditional convolution in expressing nonlinear patterns in high-dimensional data. Additionally, the dynamic grid adjustment mechanism adaptively optimizes spline function resolution by analyzing input data statistics (e.g., distribution of sorted activation values), specifically updating grid points as $\mathbf{t}_i = \epsilon \mathbf{t}_{\text{uniform}} + (1-\epsilon) \mathbf{t}_{\text{adaptive}}$, where $\epsilon$ controls the blending ratio between uniform and adaptive grids. This ensures efficient capture of detailed spectral variations while significantly reducing overfitting risks, especially on small-sample or high-noise hyperspectral datasets.Benefiting from the theoretical support of the Kolmogorov-Arnold representation theorem, this module effectively alleviates the curse of dimensionality in hyperspectral data by decomposing high-dimensional functions into combinations of univariate functions, demonstrating superior neural scaling laws ($\ell \propto N^{-4}$) compared to traditional convolutional neural networks. Without increasing network depth or width, this method enhances spatial feature representation through attention mechanisms, providing fine-grained discriminative information for classification tasks. KANet can reduce the aggregation of redundant spectral information, avoiding the uniform sampling and dimensionality reduction approach of traditional 3D-CNNs across all spectral dimensions, thereby effectively mitigating parameter redundancy and the curse of dimensionality caused by excessive spectral dimensions.

The main contributions of this paper are as follows:

1. This paper proposes a novel three-dimensional Kolmogorov-Arnold convolution module, improving the efficient 3D-DenseNet-based spectral-spatial joint hyperspectral image classification method, including 3D KAN CONV and an adaptive network update mechanism. It addresses the overfitting risks caused by spatial information redundancy and the curse of dimensionality in hyperspectral data, enhancing network generalization capability. By combining dense connections with dynamic convolution generation - where dense connections facilitate feature reuse in the network - the designed 3D-DenseNet model achieves good accuracy on both IN and UP datasets.

2. This paper introduces 3D KAN Conv in 3D-CNN, incorporating parameterized univariate sample functions on the edges of convolution operations to preserve the spatial locality of convolution while enhancing spectral-spatial utilization through KAN's learnable activation functions. Secondly, through KAN's dynamic grid update mechanism, it achieves adaptive modeling of complex distributions by dynamically adjusting the grid points of sample functions based on input data statistical characteristics. These two modules jointly improve the efficient feature extraction of convolution.

3. KANet is more concise than networks combining various DL mechanisms, without complex connections and concatenations, requiring less computation. Without increasing network depth or width, it improves model representation capability through wavelet convolution with expanded receptive fields.

\section{3D KAN Convolution with Grid Update Mechanism for Classification}
\subsection{Adaptive Convolution Design}

Given the sparse and finely clustered characteristics of hyperspectral ground objects, most methods consider how to enhance the model's feature extraction capability regarding spatial-spectral dimensions, making adaptive convolution an excellent approach to strengthen spatial-spectral dimension information representation at the convolutional kernel level. LGCNet\cite{li2025spatial} designed a learnable grouped convolution structure where both input channels and convolution kernel groups can be learned end-to-end through the network. DGCNet\cite{li2023dgcnet} designed dynamic grouped convolution, introducing small feature selectors for each group to dynamically determine which part of input channels to connect based on activations of all input channels. DACNet\cite{li2025efficient} designed dynamic attention convolution using SE to generate weights and multiple parallel convolutional kernels instead of single convolution. SG-DSCNet\cite{li2025spatial_geometry} designed a Spatial-Geometry Enhanced 3D Dynamic Snake Convolutional Neural Network, introducing deformable offsets in 3D convolution to increase kernel flexibility through constrained self-learning processes, thereby enhancing the network's regional perception of ground objects and proposing multiview feature fusion. WCNet\cite{li20253d} designed a 3D convolutional network integrating wavelet transform, introducing wavelet transform to expand the receptive field of convolution through wavelet convolution, guiding CNN to better respond to low frequencies through cascading. Each convolution focuses on different frequency bands of the input signal with gradually increasing scope. EKGNet\cite{li2025expert} includes a context-associated mapping network and dynamic kernel generation module, where the context-associated mapping module translates global contextual information of hyperspectral inputs into instructions for combining base convolutional kernels, while dynamic kernels are composed of K groups of base convolutions, analogous to K different types of experts specializing in fundamental patterns across various dimensions. These methods are all based on the design concept of adaptive convolution, attempting to address the complexity of hyperspectral data in spatial-spectral dimensions through adaptive convolution mechanisms. For the information redundancy introduced by 3D-CNN in spatial-spectral dimensions, this paper designs a more reasonable KAN Conv from the fundamental characteristics of hyperspectral data, embedding KAN's function learning capability into a 3D operator that maintains the locality of convolution operations and (some form of) parameter sharing characteristics, making it particularly suitable for processing three-dimensional data like hyperspectral images (two spatial dimensions + one spectral dimension). This represents an important extension of the original KAN concept in the field of structured data processing. Secondly, through KAN's dynamic grid update mechanism, adaptive modeling of complex distributions is achieved by dynamically adjusting the grid points of sample functions based on input data statistical characteristics, fundamentally solving the problem of ground object recognition loss.

\begin{figure}[h]
\centering
\includegraphics[width=0.9\linewidth]{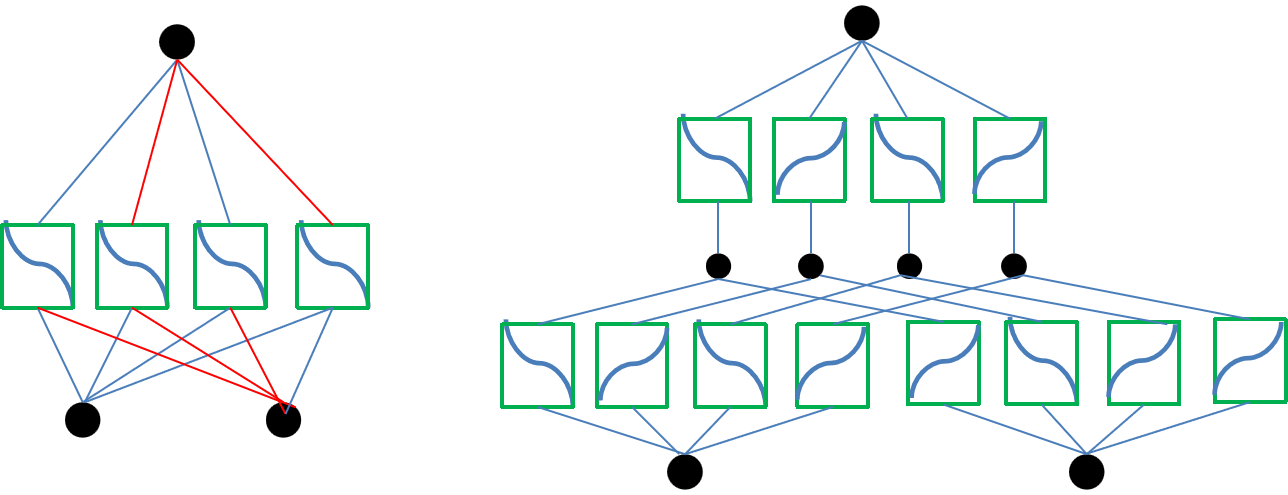}
\caption{Left: MLP structure; Right: Original KAN structure}
\label{fig:kan_structure}
\end{figure}

\subsection{KAN Convolution with Adaptive Grid Updating Mechanism}

Hyperspectral image sample data is scarce and exhibits sparse ground object characteristics, with uneven spatial distribution and substantial redundant information in the spectral dimension. Although 3D-CNN structures can utilize joint spatial-spectral information, how to more effectively achieve deep extraction of spatial-spectral information remains a noteworthy issue. As the core of convolutional neural networks, convolution kernels are generally regarded as information aggregators that combine spatial information and feature dimension information in local receptive fields. Convolutional neural networks consist of a series of convolutional layers, nonlinear layers, and downsampling layers, enabling them to capture image features from a global receptive field for image description. However, training a high-performance network is challenging, and much work has been done to improve network performance from the spatial dimension perspective. For example, the Residual structure achieves deep network extraction by fusing features produced by different blocks, while DenseNet enhances feature reuse through dense connections. 3D-CNN contains numerous redundant weights in feature extraction through convolution operations that simultaneously process spatial and spectral information of hyperspectral images. This redundancy is particularly prominent in joint spatial-spectral feature extraction: from the spatial dimension, ground objects in hyperspectral images are sparsely and unevenly distributed, and convolutional kernels may capture many irrelevant or low-information regions within local receptive fields; from the spectral dimension, hyperspectral data typically contains hundreds of bands with high correlation and redundancy between adjacent bands, making weight allocation of convolutional kernels along the spectral axis difficult to effectively focus on key features. Especially the redundant characteristics of spectral dimensions cause many convolutional parameters to only serve as "fillers" in high-dimensional data without fully mining deep patterns in joint spatial-spectral information. This weight redundancy not only increases computational complexity but may also weaken the model's representation capability for sparse ground objects and complex spectral features, thereby limiting 3D-CNN's performance in hyperspectral image processing.

KANet proposes a groundbreaking three-dimensional Kolmogorov-Arnold convolutional module that introduces learnable univariate B-spline functions on network edges to completely replace the fixed linear weight matrices of traditional 3D convolutional kernels, significantly enhancing the modeling capability for complex nonlinear spectral-spatial relationships in hyperspectral data. Through a dynamic adaptive grid optimization mechanism, it precisely adjusts the resolution of spline functions to efficiently adapt to the non-uniform distribution characteristics of high-dimensional data. The core innovation lies in reshaping three-dimensional neighborhood data into high-dimensional vectors and then applying nonlinear activation functions parameterized by B-splines, transforming the linear weighting operations of traditional convolution into spline-based nonlinear mappings, where each activation function is defined by trainable spline coefficients and basis functions (such as SiLU activation), with independent scaling factors further optimizing their expressive capabilities. This design directly draws from the theoretical essence of the Kolmogorov-Arnold representation theorem, effectively avoiding the curse of dimensionality caused by the high-dimensional nature of hyperspectral data by decomposing high-dimensional functions into combinations of univariate functions. Compared to traditional convolution's fixed parameter patterns (such as linear transformations based on fixed weights $y = W * x + b$), this module not only preserves the spatial locality of convolution operations but also achieves fine modeling of high-dimensional spectral features through learnable nonlinear functions $\phi_i$, overcoming the limitations of traditional convolution in expressing nonlinear patterns in high-dimensional data. Additionally, the dynamic grid adjustment mechanism analyzes the statistical characteristics of input data (such as the distribution of sorted activation values) to adaptively optimize the resolution of spline functions, specifically manifested in updating grid points $t_i = \epsilon t_{\text{uniform}} + (1-\epsilon) t_{\text{adaptive}}$, where $\epsilon$ controls the fusion ratio between uniform and adaptive grids. This ensures the model can efficiently capture subtle changes in the spectral dimension, particularly significantly reducing overfitting risks in small-sample or high-noise hyperspectral datasets. Benefiting from the theoretical support of the Kolmogorov-Arnold representation theorem, this module effectively alleviates the curse of dimensionality caused by the high-dimensional nature of hyperspectral data by decomposing high-dimensional functions into combinations of univariate functions, demonstrating superior neural scaling laws ($\ell \propto N^{-4}$) compared to traditional convolutional neural networks. This method enhances spatial feature representation capability through attention mechanisms without increasing network depth or width. The high-dimensional spectral information provides fine-grained discrimination basis for classification tasks. KANet can reduce aggregation of redundant spectral information, avoiding the uniform sampling and dimensionality reduction treatment of all spectral dimensions in traditional 3D-CNN, effectively alleviating parameter redundancy and computational complexity caused by excessive spectral dimensions. Through the deep characteristics of 3D-DenseNet combined with dynamic convolution generation, features can be extracted more effectively.

\begin{figure}[h]
\centering
\includegraphics[width=0.9\linewidth]{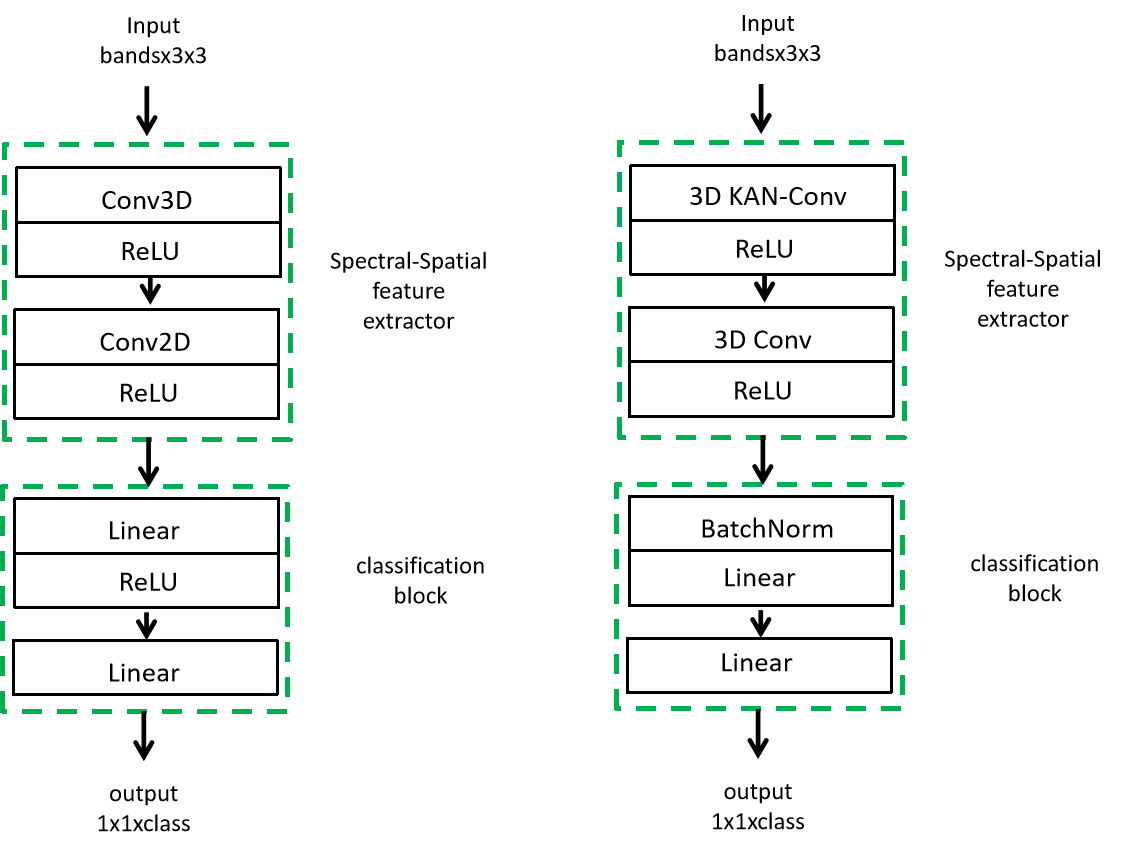}
\caption{Left: Standard 3D-CNN structure; Right: 3D-KAN Conv structure}
\label{fig:3d_conv_comparison}
\end{figure}

\subsubsection{3D KAN Conv}
KANet introduces a novel three-dimensional Kolmogorov-Arnold convolutional module that deploys learnable univariate B-spline functions on network edges to completely replace the fixed linear weight matrices of traditional 3D convolution, greatly enhancing nonlinear modeling capabilities for complex high-dimensional data and providing an efficient convolutional paradigm for data-intensive tasks. The core design flattens the three-dimensional neighborhood of input data (defined by \texttt{kernel\_size}) into high-dimensional vectors through \texttt{unfoldNd} operations, generating tensors of shape $[B, N, C\cdot K_1\cdot K_2\cdot K_3]$, where $B$ is batch size, $N$ is the number of output spatial positions, and $C\cdot K_1\cdot K_2\cdot K_3$ is the flattened feature dimension (implemented in \texttt{effConvKAN3D.forward}). Subsequently, the \texttt{KANLinear} layer applies B-spline-parameterized nonlinear activation functions, defined as $\phi(x) = w_b b(x) + w_s \sum_i c_i B_i(x)$, where $b(x) = \text{SiLU}(x)$ is the basis function, $B_i(x)$ are B-spline basis functions (implemented in \texttt{b\_splines}), $c_i$ are trainable coefficients (\texttt{spline\_weight}), and $w_b$ and $w_s$ are scaling factors. This nonlinear mapping replaces the linear transformation $\mathbf{y} = \mathbf{W} * \mathbf{x} + \mathbf{b}$ of traditional convolution with $\mathbf{y} = \sum_i \phi_i(\mathbf{x}_i;\mathbf{c}_i,\mathbf{t}_i)$, significantly improving the expressive capability for complex nonlinear patterns in high-dimensional features. B-spline basis functions are calculated through recursive formulas (loops in \texttt{b\_splines}), specifically:

$$B_i^k(x) = \frac{x - t_i}{t_{i+k} - t_i} B_i^{k-1}(x) + \frac{t_{i+k+1} - x}{t_{i+k+1} - t_{i+1}} B_{i+1}^{k-1}(x),$$

where $t_i$ are grid points, $k$ is the spline order . The coefficients $c_i$ are determined through least squares optimization $\min_c ||Ac - y||_2^2$, where $A$ is the B-spline basis matrix . Compared to the fixed weights of traditional convolution, 3D KAN Conv achieves higher parameter efficiency through the local support and learnability of B-splines (initialized by \texttt{reset\_parameters}). Experiments show it achieves excellent performance in classification tasks with fewer parameters, validating the superiority of neural scaling laws in KAN theory. Through parameters like \texttt{stride}, \texttt{padding}, and \texttt{dilation}, this module preserves the spatial locality of convolution, ensuring effective modeling of spatial structures while providing potential interpretability for feature analysis.

\subsubsection{Dynamic Grid Optimization Mechanism}
KANet enhances the performance of the three-dimensional Kolmogorov-Arnold convolutional module through a dynamic grid optimization mechanism that adaptively adjusts the positions of B-spline grid points based on input data statistical characteristics, optimizing the resolution of activation functions $\phi_i(x;c_i,t_i)$ to precisely match the non-uniform distribution of data, significantly improving model accuracy and robustness, providing an adaptive convolutional framework for high-dimensional data processing. This mechanism analyzes the distribution of input activation values to generate adaptive grids $t_{\text{adaptive}}$, calculated based on equally spaced sampling of sorted activation values:

$$t_{\text{adaptive}} = \text{sort}(\mathbf{x})[\text{linspace}(0, B-1, G+1)],$$

where $B$ is batch size and $G$ is the number of grid points (\texttt{grid\_size}). Subsequently, it merges with uniform grids $t_{\text{uniform}}$, defined as:

$$t_{\text{uniform}} = \left[\text{range}(0, G+1) \cdot \frac{\max(\mathbf{x}) - \min(\mathbf{x}) + 2m}{G} + \min(\mathbf{x}) - m\right],$$

where $m$ is the boundary margin (\texttt{margin}). The final grid is generated through weighted fusion:

$$t_i = \epsilon t_{\text{uniform}} + (1-\epsilon) t_{\text{adaptive}},$$

where $\epsilon$ is the fusion ratio. To maintain the continuity of spline functions, the grid is extended to include boundary points, with the updated grid stored in the grid buffer. Spline coefficients $c_i$ are recalculated through least squares optimization:

$$c_i = \text{lstsq}(A(x, t_i), y),$$

where $A$ is the updated B-spline basis matrix. This process ensures activation functions dynamically adapt to data distribution, finely capturing feature changes, particularly reducing overfitting risks in high-noise scenarios. Compared to fixed parameters in traditional convolution, this mechanism significantly improves modeling capability for complex nonlinear relationships by optimizing the approximation accuracy of univariate functions. Experiments show it achieves higher performance and parameter efficiency in classification tasks, validating the superiority of neural scaling laws $\ell \propto N^{-4}$ in KAN theory. In hyperspectral applications, this mechanism provides flexible modeling tools for data analysis by dynamically adjusting grids to adapt to changes in ground object spectral features.

This characteristic enables KANet to efficiently capture subtle changes in the spectral dimension, particularly significantly reducing overfitting risks in small-sample or high-noise scenarios. Compared to traditional 3D convolution, this module not only preserves the spatial locality of convolution operations but also achieves higher parameter utilization efficiency through the local support and learnability of B-splines. Experimental verification shows it achieves excellent classification performance in hyperspectral classification tasks with fewer parameters, fully demonstrating the superiority of neural scaling laws in KAN theory. By integrating this mechanism into the 3D-DenseNet framework, KANet achieves enhanced feature reuse and representation capability without introducing excessive parameters. The synergy between the context-aware mapping network and expert convolution system ensures KANet effectively mitigates challenges posed by high-dimensional redundancy and sparse ground object distribution, achieving exceptional classification performance.

\begin{figure}[h]
\centering
\includegraphics[width=0.9\linewidth]{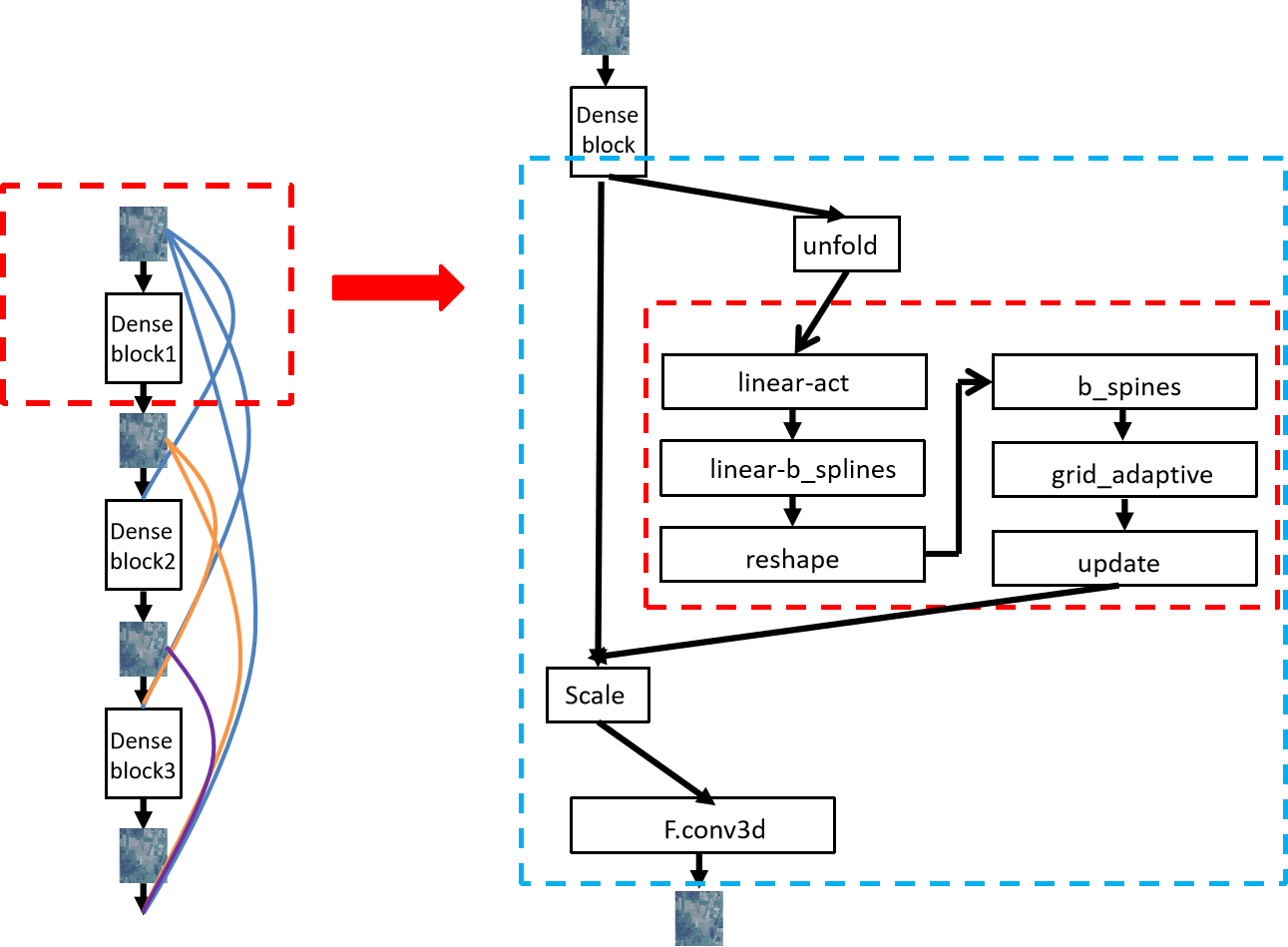}
\caption{Design of KANet in 3D-DenseNet's dense block}
\label{fig:dense_block_design}
\end{figure}

\subsection{3D-CNN Framework for Hyperspectral Image Feature Extraction and Model Implementation}

We made two modifications to the original 3D-DenseNet to further simplify the architecture and improve its computational efficiency.

\subsubsection{Exponentially Increasing Growth Rate}
The original DenseNet design adds $k$ new feature maps per layer, where $k$ is a constant called the growth rate. As shown in \cite{huang2017densely}, deeper layers in DenseNet tend to rely more on high-level features than low-level ones, which motivated our improvement through strengthened short connections. We found this could be achieved by gradually increasing the growth rate with depth. This increases the proportion of features from later layers relative to earlier ones. For simplicity, we set the growth rate as $k=2^{m-1}k_0$, where $m$ is the dense block index and $k_0$ is a constant. This growth rate setting does not introduce any additional hyperparameters. The "increasing growth rate" strategy places a larger proportion of parameters in the model's later layers. This significantly improves computational efficiency, though it may reduce parameter efficiency in some cases. Depending on specific hardware constraints, trading one for the other may be advantageous \cite{liu2017learning}.

\begin{figure}[h]
\centering
\includegraphics[width=0.9\linewidth]{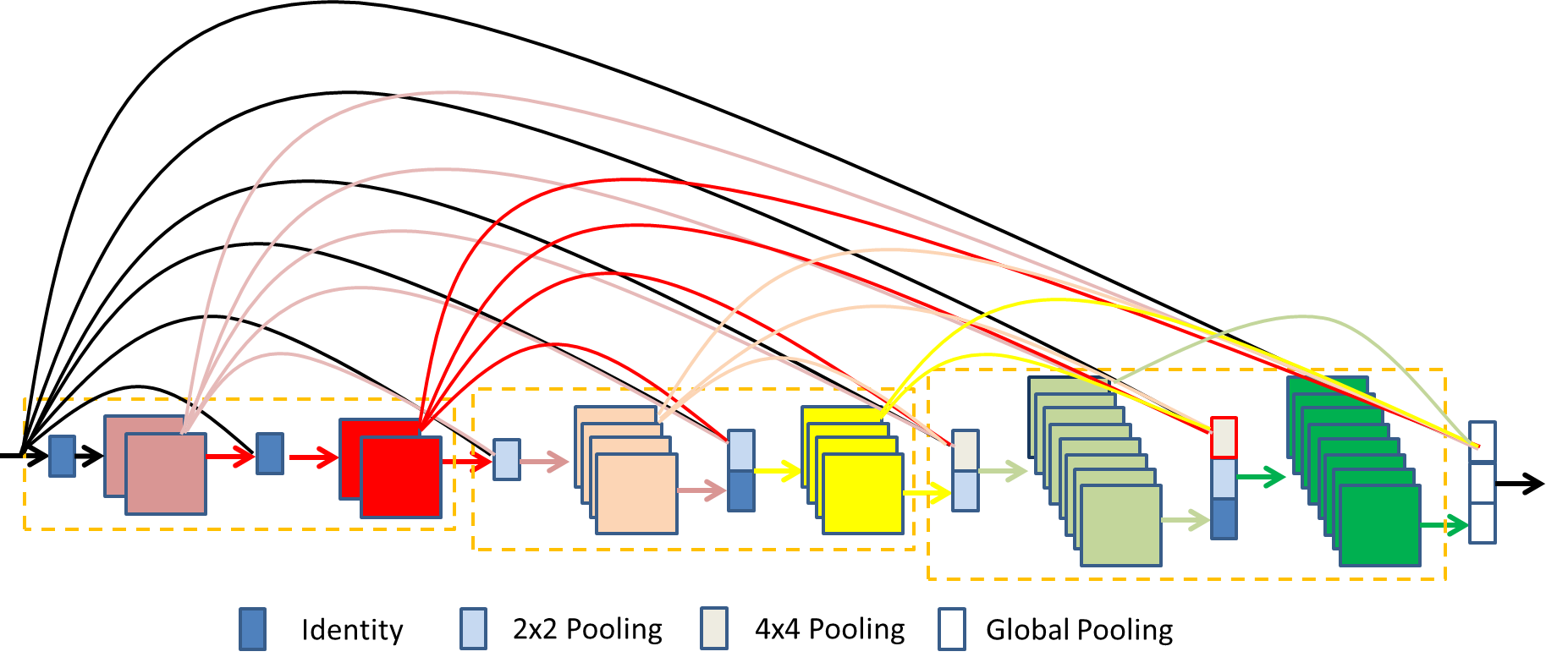}
\caption{Proposed DenseNet variant with two key differences from original DenseNet: (1) Direct connections between layers with different feature resolutions; (2) Growth rate doubles when feature map size reduces (third yellow dense block generates significantly more features than the first block)}
\label{fig:densenet_variant}
\end{figure}

\subsubsection{Fully Dense Connectivity}
To encourage greater feature reuse than the original DenseNet architecture, we connect the input layer to all subsequent layers across different dense blocks (see Figure~\ref{fig:densenet_variant}). Since dense blocks have different feature resolutions, we downsample higher-resolution feature maps using average pooling when connecting them to lower-resolution layers.

The overall model architecture is shown below:

\begin{figure}[h]
\centering
\includegraphics[width=0.9\linewidth]{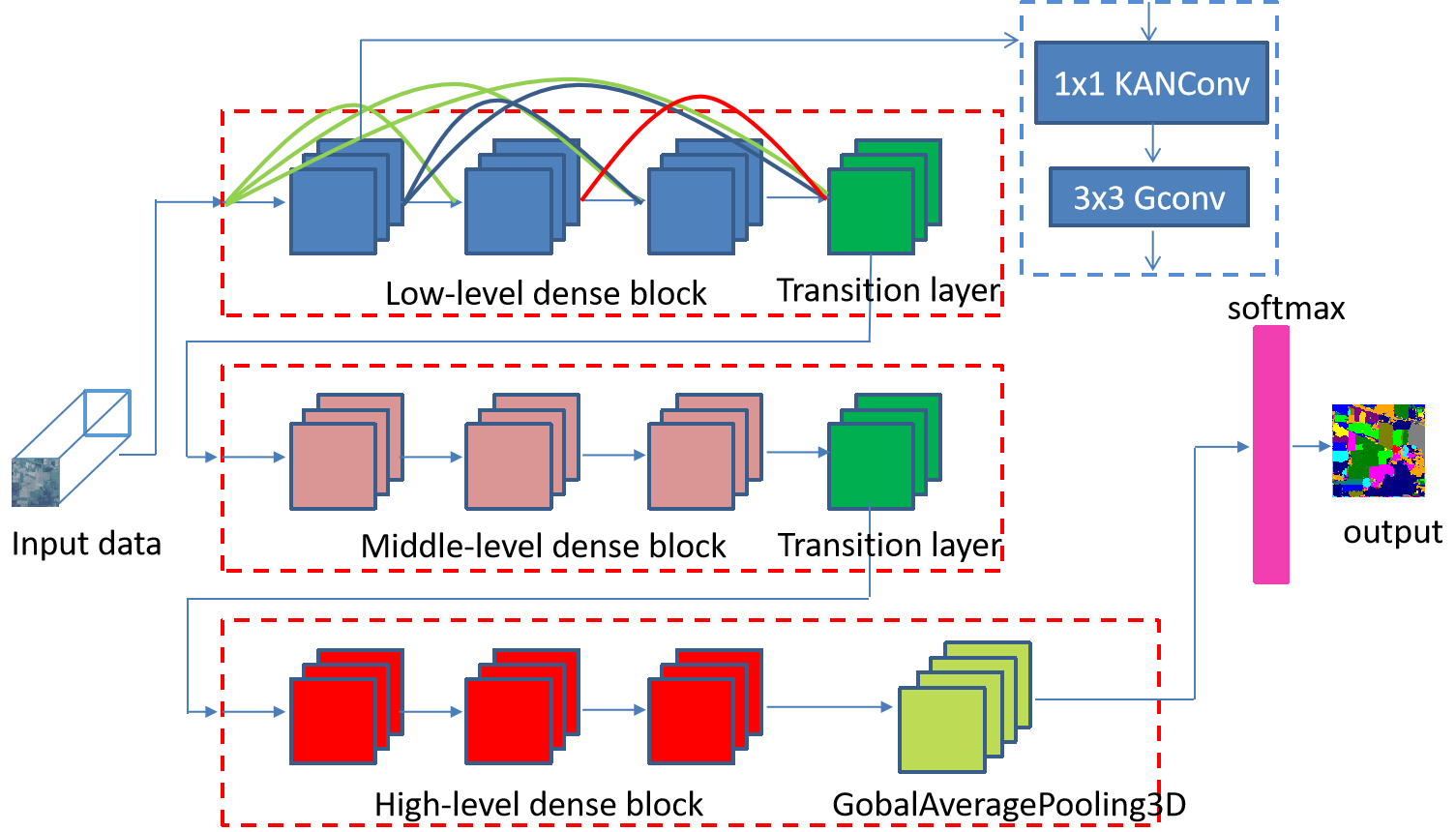}
\caption{Overall architecture of our KANet, incorporating the 3D-DenseNet basic framework}
\label{fig:kanet_architecture}
\end{figure}

\section{Experiments and Analysis}

To evaluate the performance of KANet, we conducted experiments on three representative hyperspectral datasets: Indian Pines, Pavia University, and Kennedy Space Center (KSC). The classification metrics include Overall Accuracy (OA), Average Accuracy (AA), and Kappa coefficient.

\subsection{Experimental Datasets}
\subsubsection{Indian Pines Dataset}
\begin{figure}[h]
\centering
\includegraphics[width=0.9\linewidth]{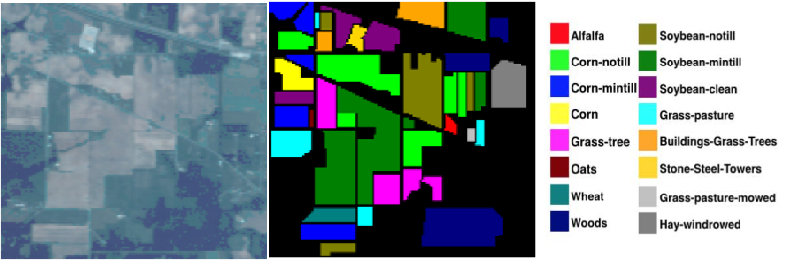}
\caption{False color composite and ground truth labels of Indian Pines dataset}
\label{fig:indian_pines}
\end{figure}

The Indian Pines dataset was collected in June 1992 by the AVIRIS (Airborne Visible/Infrared Imaging Spectrometer) sensor over a pine forest test site in northwestern Indiana, USA. The dataset consists of $145\times145$ pixel images with a spatial resolution of 20 meters, containing 220 spectral bands covering the wavelength range of 0.4--2.5$\mu$m. In our experiments, we excluded 20 bands affected by water vapor absorption and low signal-to-noise ratio (SNR), utilizing the remaining 200 bands for analysis. The dataset encompasses 16 land cover categories including grasslands, buildings, and various crop types. Figure~\ref{fig:indian_pines} displays the false-color composite image and spatial distribution of ground truth samples.

\subsubsection{Pavia University Dataset}
\begin{figure}[h]
\centering
\includegraphics[width=0.9\linewidth]{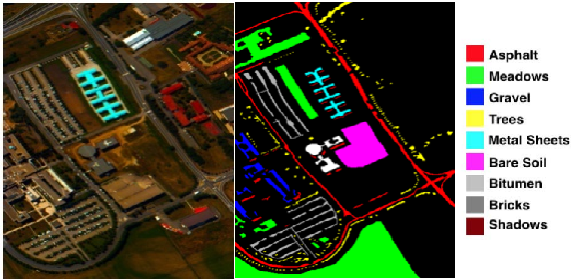}
\caption{False color composite and ground truth labels of Pavia University dataset}
\label{fig:pavia_university}
\end{figure}

The Pavia University dataset was acquired in 2001 by the ROSIS imaging spectrometer over the Pavia region in northern Italy. The dataset contains images of size $610\times340$ pixels with a spatial resolution of 1.3 meters, comprising 115 spectral bands in the wavelength range of 0.43--0.86$\mu$m. For our experiments, we removed 12 bands containing strong noise and water vapor absorption, retaining 103 bands for analysis. The dataset includes 9 land cover categories such as roads, trees, and roofs. Figure~\ref{fig:pavia_university} shows the spatial distribution of different classes.

\subsubsection{Kennedy Space Center Dataset}
The KSC dataset was collected on March 23, 1996 by the AVIRIS imaging spectrometer over the Kennedy Space Center in Florida. The AVIRIS sensor captured 224 spectral bands with 10 nm width, centered at wavelengths from 400 to 2500 nm. Acquired from an altitude of approximately 20 km, the dataset has a spatial resolution of 18 meters. After removing bands affected by water absorption and low signal-to-noise ratio (SNR), we used 176 bands for analysis, which represent 13 defined land cover categories.

\subsection{Experimental Analysis}
KANet was trained for 80 epochs on all three datasets using the Adam optimizer. The experiments were conducted on a platform with four 80GB A100 GPUs. For analysis, we employed the kanet-base architecture with three stages, where each stage contained 4, 6, and 8 dense blocks respectively. The growth rates were set to 8, 16, and 32, with 4 heads. The 3×3 group convolution used 4 groups, gate factor was 0.25, and the compression ratio was 16.

\subsubsection{Data Partitioning Ratio}
For hyperspectral data with limited samples, the training set ratio significantly impacts model performance. To systematically evaluate the sensitivity of data partitioning strategies, we compared the model's generalization performance under different Train/Validation/Test ratios. Experiments show that with limited training samples, a 6:1:3 ratio effectively balances learning capability and evaluation reliability on Indian Pines - this configuration allocates 60\% samples for training, 10\% for validation (enabling early stopping to prevent overfitting), and 30\% for statistically significant testing. KANet adopted 4:1:5 ratio on Pavia University and 5:1:4 ratio on KSC datasets, with 11×11 neighboring pixel blocks to balance local feature extraction and spatial context integrity.

\begin{table}[h]
\begin{minipage}{\textwidth}
\centering
\makeatletter
\def\@makecaption#1#2{%
    \vskip\abovecaptionskip
    \centering 
    \small #1: #2\par
    \vskip\belowcaptionskip
}
\makeatother
\caption{OA, AA and Kappa metrics for different training set ratios on the Indian Pines dataset}
\begin{tabular}{cccc}
\toprule
Training Ratio & OA & AA & Kappa \\
\midrule
2:1:7 & 92.12 & 93.05 & 91.00 \\
3:1:6 & 96.96 & 95.27 & 96.53 \\
4:1:5 & 99.10 & 99.29 & 98.98 \\
5:1:4 & 98.97 & 98.34 & 98.83 \\
6:1:3 & 99.84 & 99.65 & 99.81 \\
\bottomrule
\end{tabular}
\label{tab:indian_ratios}
    \end{minipage}

 \vspace*{10pt} 

\begin{minipage}{\textwidth}
\centering
\makeatletter
\def\@makecaption#1#2{%
    \vskip\abovecaptionskip
    \centering 
    \small #1: #2\par
    \vskip\belowcaptionskip
}
\makeatother
\caption{OA, AA and Kappa metrics for different training set ratios on the Pavia University dataset}
\begin{tabular}{cccc}
\toprule
Training Ratio & OA & AA & Kappa \\
\midrule
2:1:7 & 99.46 & 99.37 & 99.29 \\
3:1:6 & 99.57 & 99.44 & 99.43 \\
4:1:5 & 99.92 & 99.92 & 99.89 \\
5:1:4 & 99.92 & 99.87 & 99.90 \\
6:1:3 & 99.94 & 99.87 & 99.92 \\
\bottomrule
\end{tabular}
\label{tab:pavia_ratios}
        \end{minipage}

 \vspace*{10pt} 
 
\begin{minipage}{\textwidth} 
\centering
\makeatletter
\def\@makecaption#1#2{%
    \vskip\abovecaptionskip
    \centering 
    \small #1: #2\par
    \vskip\belowcaptionskip
}
\caption{OA, AA and Kappa metrics for different training set ratios on the KSC dataset}
\begin{tabular}{cccc}
\toprule
Training Ratio & OA & AA & Kappa \\
\midrule
2:1:7 & 97.72 & 96.67 & 97.46 \\
3:1:6 & 98.81 & 98.25 & 98.68 \\
4:1:5 & 99.31 & 99.20 & 99.23 \\
5:1:4 & 99.62 & 99.32 & 99.57 \\
6:1:3 & 99.55 & 99.47 & 99.50 \\
\bottomrule
\end{tabular}
\label{tab:ksc_ratios}
        \end{minipage}
\end{table}

\subsubsection{Neighboring Pixel Blocks}
The network performs edge padding on the input $145\times145\times103$ image (using Indian Pines as an example), transforming it into a $155\times155\times103$ image. On this padded image, it sequentially selects adjacent pixel blocks of size $M\times N\times L$, where $M\times N$ represents the spatial sampling size and $L$ is the full spectral dimension. Large original images are unfavorable for convolutional feature extraction, leading to slower processing speeds, temporary memory spikes, and higher hardware requirements. Therefore, we adopt adjacent pixel block processing. The block size is a crucial hyperparameter - too small blocks may result in insufficient receptive fields for convolutional feature extraction, leading to poor local performance. As shown in Tables 4-6, on the Indian Pines dataset, accuracy shows significant improvement when block size increases from 7 to 17. However, the accuracy gain diminishes with larger blocks, showing a clear threshold effect. At block size 17, accuracy even decreases slightly, a phenomenon also observed in Pavia University and KSC datasets. Consequently, we select block size 15 for Indian Pines and 17 for Pavia University and KSC datasets.

\begin{table*}[h]
\begin{minipage}{\textwidth}
\centering
\makeatletter
\def\@makecaption#1#2{%
    \vskip\abovecaptionskip
    \centering 
    \small #1: #2\par
    \vskip\belowcaptionskip
}
\makeatother
\caption{OA, AA and Kappa metrics for different block sizes on Indian Pines}
\begin{tabular}{cccc}
\toprule
Block Size (M=N) & OA & AA & Kappa \\
\midrule
7 & 96.94 & 96.77 & 96.51 \\
9 & 97.95 & 98.59 & 97.66 \\
11 & 99.84 & 99.65 & 99.81 \\
13 & 99.90 & 99.93 & 99.89 \\
15 & 99.94 & 99.77 & 99.93 \\
17 & 99.87 & 99.70 & 99.85 \\
\hline
\end{tabular}
        \end{minipage}

 \vspace*{10pt} 

\begin{minipage}{\textwidth}
\centering
\makeatletter
\def\@makecaption#1#2{%
    \vskip\abovecaptionskip
    \centering 
    \small #1: #2\par
    \vskip\belowcaptionskip
}
\makeatother
\caption{OA, AA and Kappa metrics for different block sizes on Pavia University}
\begin{tabular}{cccc}
\toprule
Block Size (M=N) & OA & AA & Kappa \\
\midrule
7 & 99.80 & 99.72 & 99.74 \\
9 & 99.84 & 99.77 & 99.78 \\
11 & 99.92 & 99.92 & 99.89 \\
13 & 99.97 & 99.97 & 99.96 \\
15 & 99.99 & 99.98 & 99.98 \\
17 & 99.99 & 99.99 & 99.99 \\
\hline
\end{tabular}
\end{minipage}

 \vspace*{10pt} 
 
\begin{minipage}{\textwidth}
\centering
\makeatletter
\def\@makecaption#1#2{%
    \vskip\abovecaptionskip
    \centering 
    \small #1: #2\par
    \vskip\belowcaptionskip
}
\makeatother
\caption{OA, AA and Kappa metrics for different block sizes on KSC}
\begin{tabular}{cccc}
\toprule
Block Size (M=N) & OA & AA & Kappa \\
\midrule
7 & 98.61 & 97.82 & 98.45 \\
9 & 99.52 & 99.29 & 99.46 \\
11 & 99.62 & 99.32 & 99.57 \\
13 & 99.62 & 99.45 & 99.57 \\
15 & 99.86 & 99.76 & 99.84 \\
17 & 100.00 & 100.00 & 100.00 \\
\hline
\end{tabular}
        \end{minipage}
\end{table*}

\subsection{Network Parameters}
We categorize KANet into base and large types. Table 7 shows OA, AA, and Kappa results on Indian Pines dataset. While the large model with more parameters shows improved accuracy, the gain is not substantial compared to the rapid parameter increase in dense blocks, suggesting KANet-base already possesses excellent generalization and classification capabilities.

\begin{table*}[!ht]
\centering
\makeatletter
\def\@makecaption#1#2{%
    \vskip\abovecaptionskip
    \centering 
    \small #1: #2\par
    \vskip\belowcaptionskip
}
\makeatother
\caption{KANet model configurations and performance metrics}
\begin{tabular}{cccccc}
\toprule
Model & Stages/Dense Block & Growth Rate & OA & AA & Kappa \\
\midrule
KANet-base & 4,6,8 & 8,16,32 & 99.94 & 99.77 & 99.93 \\
KANet-large & 14,14,14 & 8,16,32 & 99.96 & 99.81 & 99.92 \\
\bottomrule
\end{tabular}
\label{tab:params}
\end{table*}

\subsection{Experimental Results}
On Indian Pines, Pavia University, and KSC datasets, KANet uses input sizes of $17\times17\times200$, $17\times17\times103$, and $17\times17\times176$ respectively. We compare KANet-base with SSRN, 3D-CNN, 3D-SE-DenseNet, Spectralformer, lgcnet, and dgcnet (Tables 8-9). KANet variants consistently achieve leading accuracy. Figure 9 shows training/validation loss and accuracy curves, demonstrating rapid convergence and stable accuracy improvement.

\begin{table*}[!ht]
\centering
\makeatletter
\def\@makecaption#1#2{%
    \vskip\abovecaptionskip
    \centering 
    \small #1: #2\par
    \vskip\belowcaptionskip
}
\makeatother
\caption{Classification accuracy comparison (\%) on Indian Pines dataset}
\label{tab:indian_results}
\resizebox{\textwidth}{!}{
\begin{tabular}{@{}lccccccc@{}}
\toprule
Class & SSRN & 3D-CNN & 3D-SE-DenseNet & Spectralformer & LGCNet & DGCNet & KANet \\
\midrule
1 & 100 & 96.88 & 95.87 & 70.52 & 100 & 100 & 100 \\
2 & 99.85 & 98.02 & 98.82 & 81.89 & 99.92 & 99.47 & 100 \\
3 & 99.83 & 97.74 & 99.12 & 91.30 & 99.87 & 99.51 & 100 \\
4 & 100 & 96.89 & 94.83 & 95.53 & 100 & 97.65 & 100 \\
5 & 99.78 & 99.12 & 99.86 & 85.51 & 100 & 100 & 100 \\
6 & 99.81 & 99.41 & 99.33 & 99.32 & 99.56 & 99.88 & 100 \\
7 & 100 & 88.89 & 97.37 & 81.81 & 95.83 & 100 & 100 \\
8 & 100 & 100 & 100 & 75.48 & 100 & 100 & 100 \\
9 & 0 & 100 & 100 & 73.76 & 100 & 100 & 100 \\
10 & 100 & 100 & 99.48 & 98.77 & 99.78 & 98.85 & 100 \\
11 & 99.62 & 99.33 & 98.95 & 93.17 & 99.82 & 99.72 & 99.86 \\
12 & 99.17 & 97.67 & 95.75 & 78.48 & 100 & 99.56 & 100 \\
13 & 100 & 99.64 & 99.28 & 100 & 100 & 100 & 100 \\
14 & 98.87 & 99.65 & 99.55 & 79.49 & 100 & 99.87 & 100 \\
15 & 100 & 96.34 & 98.70 & 100 & 100 & 100 & 100 \\
16 & 98.51 & 97.92 & 96.51 & 100 & 97.73 & 98.30 & 96.43 \\
\midrule
OA & 99.62±0.00 & 98.23±0.12 & 98.84±0.18 & 81.76 & 99.85±0.04 & 99.58 & 99.94 \\
AA & 93.46±0.50 & 98.80±0.11 & 98.42±0.56 & 87.81 & 99.53±0.23 & 99.55 & 99.77 \\
K & 99.57±0.00 & 97.96±0.53 & 98.60±0.16 & 79.19 & 99.83±0.05 & 99.53 & 99.93 \\
\bottomrule
\end{tabular}
}
\end{table*}

\begin{figure*}[!ht]
\centering
\includegraphics[width=0.8\linewidth]{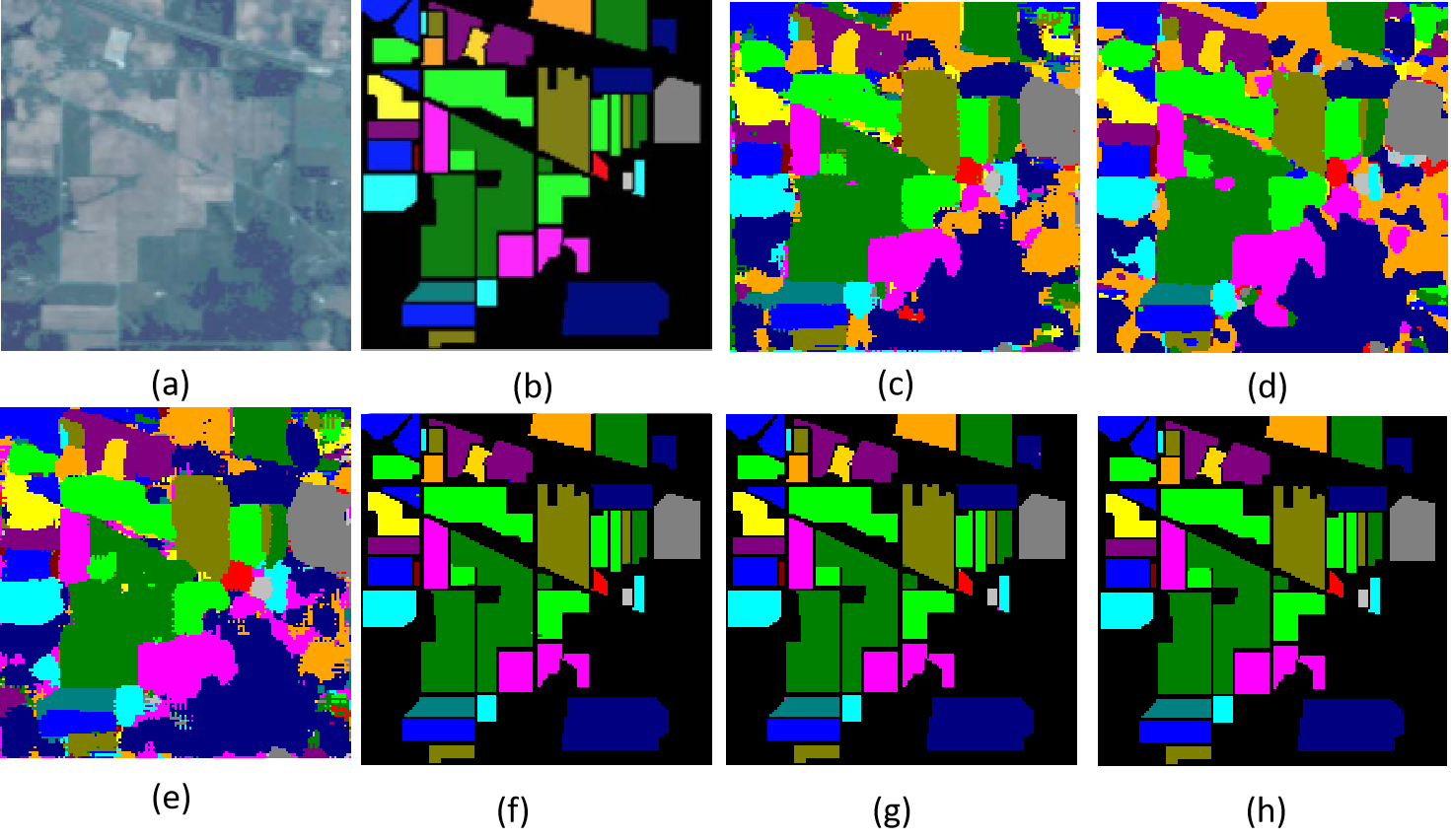}
\caption{Classification results comparison for Indian Pines dataset: (a) False color image, (b) Ground-truth labels, (c)-(i) Classification results of SSRN, 3D-CNN, 3D-SE-DenseNet, LGCNet, DGCNet, and KANet}
\label{fig:classification}
\end{figure*}

\begin{figure*}[ht]
\centering
\includegraphics[width=0.8\linewidth]{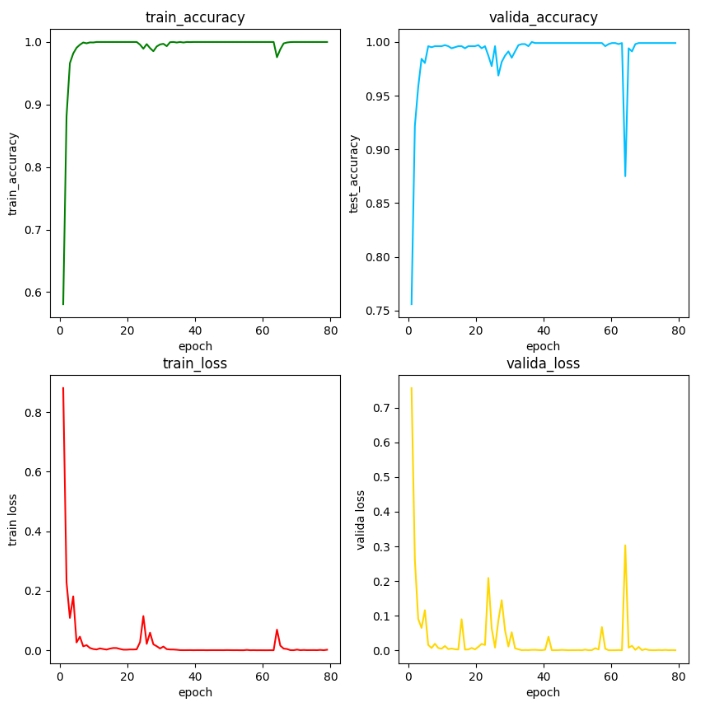}
\caption{Training and validation curves of KANet showing loss and accuracy evolution}
\label{fig:training_curve}
\end{figure*}

\begin{table*}[!ht]
\centering
\makeatletter
\def\@makecaption#1#2{%
    \vskip\abovecaptionskip
    \centering 
    \small #1: #2\par
    \vskip\belowcaptionskip
}
\makeatother
\caption{Classification accuracy (\%) comparison on Pavia University dataset}
\label{tab:pavia_results}
\resizebox{\textwidth}{!}{
\begin{tabular}{@{}lcccccc@{}}
\toprule
Class & SSRN & 3D-CNN & 3D-SE-DenseNet & Spectralformer & LGCNet & KANet \\
\midrule
1 & 89.93 & 99.96 & 99.32 & 82.73 & 100 & 100 \\
2 & 86.48 & 99.99 & 99.87 & 94.03 & 100 & 100 \\
3 & 99.95 & 99.64 & 96.76 & 73.66 & 99.88 & 100 \\
4 & 95.78 & 99.83 & 99.23 & 93.75 & 100 & 100 \\
5 & 97.69 & 99.81 & 99.64 & 99.28 & 100 & 100 \\
6 & 95.44 & 99.98 & 99.80 & 90.75 & 100 & 99.97 \\
7 & 84.40 & 97.97 & 99.47 & 87.56 & 100 & 100 \\
8 & 100 & 99.56 & 99.32 & 95.81 & 100 & 100 \\
9 & 87.24 & 100 & 100 & 94.21 & 100 & 100 \\
\midrule
OA & 92.99±0.39 & 99.79±0.01 & 99.48±0.02 & 91.07 & 99.99±0.00 & 99.99 \\
AA & 87.21±0.25 & 99.75±0.15 & 99.16±0.37 & 90.20 & 99.99±0.01 & 99.99 \\
K & 90.58±0.18 & 99.87±0.27 & 99.31±0.03 & 88.05 & 99.99±0.00 & 99.99 \\
\bottomrule
\end{tabular}
}
\end{table*}

\section{Conclusion}
This paper proposes a novel KANet designed for 3D convolutional kernels, incorporating 3D KAN convolution and an adaptive grid update mechanism. By introducing learnable univariate B-spline functions at the network's edges, the approach flattens the 3D neighborhood into a vector and applies nonlinear activation functions parameterized by B-splines, replacing the fixed linear weights of traditional 3D convolutional kernels. This enables precise capture of complex spectral-spatial nonlinear relationships in hyperspectral data. Additionally, a dynamic grid adjustment mechanism adaptively updates the grid point positions of B-splines based on the statistical characteristics of the input data, optimizing the spline function's resolution to match the non-uniform distribution of spectral features. This significantly enhances the model's modeling accuracy and parameter efficiency for high-dimensional data, effectively mitigating the curse of dimensionality. These characteristics demonstrate superior neural scaling laws compared to traditional convolutional neural networks, reducing overfitting risks in small-sample and high-noise scenarios. The approach further improves the model's ability to represent joint spatial-spectral information, effectively skipping redundant information and reducing computational complexity. KANet leverages the 3D-DenseNet architecture to extract critical spatial structures and spectral information, providing an efficient solution for hyperspectral image classification. It successfully addresses challenges posed by sparse ground object distributions and spectral redundancy.

\section*{Data Availability Statement}
The datasets used in this study are publicly available and widely used benchmark datasets in the hyperspectral image analysis community.

{\small
\bibliographystyle{template}
\bibliography{template}
}

\end{document}